# Cached Sufficient Statistics for Efficient Machine Learning with Large Datasets

**Andrew Moore**                                        AWM@CS.CMU.EDU
**Mary Soon Lee**                                       MSLEE@CS.CMU.EDU
*School of Computer Science and Robotics Institute*
*Carnegie Mellon University, Pittsburgh PA 15213*

## Abstract

This paper introduces new algorithms and data structures for quick counting for machine learning datasets. We focus on the counting task of constructing contingency tables, but our approach is also applicable to counting the number of records in a dataset that match conjunctive queries. Subject to certain assumptions, the costs of these operations can be shown to be independent of the number of records in the dataset and loglinear in the number of non-zero entries in the contingency table.

We provide a very sparse data structure, the *AD*tree, to minimize memory use. We provide analytical worst-case bounds for this structure for several models of data distribution. We empirically demonstrate that tractably-sized data structures can be produced for large real-world datasets by (a) using a sparse tree structure that never allocates memory for counts of zero, (b) never allocating memory for counts that can be deduced from other counts, and (c) not bothering to expand the tree fully near its leaves.

We show how the *AD*tree can be used to accelerate Bayes net structure finding algorithms, rule learning algorithms, and feature selection algorithms, and we provide a number of empirical results comparing *AD*tree methods against traditional direct counting approaches. We also discuss the possible uses of *AD*trees in other machine learning methods, and discuss the merits of *AD*trees in comparison with alternative representations such as *k*d-trees, *R*-trees and Frequent Sets.

## 1. Caching Sufficient Statistics

Computational efficiency is an important concern for machine learning algorithms, especially when applied to large datasets (Fayyad, Mannila, & Piatetsky-Shapiro, 1997; Fayyad & Uthurusamy, 1996) or in real-time scenarios. In earlier work we showed how *k*d-trees with multiresolution cached regression matrix statistics can enable very fast locally weighted and instance based regression (Moore, Schneider, & Deng, 1997). In this paper, we attempt to accelerate predictions for symbolic attributes using a kind of *k*d-tree that splits on all dimensions at all nodes.

Many machine learning algorithms operating on datasets of symbolic attributes need to do frequent *counting*. This work is also applicable to Online Analytical Processing (OLAP) applications in data mining, where operations on large datasets such as multidimensional database access, DataCube operations (Harinarayan, Rajaraman, & Ullman, 1996), and association rule learning (Agrawal, Mannila, Srikant, Toivonen, & Verkamo, 1996) could be accelerated by fast counting.

Let us begin by establishing some notation. We are given a data set with $R$ records and $M$ attributes. The attributes are called $a_1, a_2, \ldots a_M$. The value of attribute $a_i$ in the $k$th record is a small integer lying in the range $\{1, 2, \ldots n_i\}$ where $n_i$ is called the *arity* of attribute $i$. Figure 1 gives an example.

Figure 1: A simple dataset used as an example. It has $R = 6$ records and $M = 3$ attributes.

## 1.1 Queries

A *query* is a set of ($attribute = value$) pairs in which the left hand sides of the pairs form a subset of $\{a_1 \ldots a_M\}$ arranged in increasing order of index. Four examples of queries for our dataset are

$$(a_1 = 1); (a_2 = 3, a_3 = 1); (); (a_1 = 2, a_2 = 4, a_3 = 2) \tag{1}$$

Notice that the total number of possible queries is $\Pi_{i=1}^{M}(n_i + 1)$. This is because each attribute can either appear in the query with one of the $n_i$ values it may take, or it may be omitted (which is equivalent to giving it a $a_i = *$ "don't care" value).

## 1.2 Counts

The *count* of a query, denoted by $C(Query)$ is simply the number of records in the dataset matching all the ($attribute = value$) pairs in $Query$. For our example dataset we find:

$$
\begin{aligned}
C(a_1 = 1) &= 3 \\
C(a_2 = 3, a_3 = 1) &= 4 \\
C() &= 6 \\
C(a_1 = 2, a_2 = 4, a_3 = 2) &= 1
\end{aligned}
$$

## 1.3 Contingency Tables

Each subset of attributes, $a_{i(1)} \ldots a_{i(n)}$, has an associated *contingency table* denoted by $\mathbf{ct}(a_{i(1)} \ldots a_{i(n)})$. This is a table with a row for each of the possible sets of values for $a_{i(1)} \ldots a_{i(n)}$. The row corresponding to $a_{i(1)} = v_1 \ldots a_{i(n)} = v_n$ records the count $C(a_{i(1)} = v_1 \ldots a_{i(n)} = v_n)$. Our example dataset has 3 attributes and so $2^3 = 8$ contingency tables exist, depicted in Figure 2.

Figure 2: The eight possible contingency tables for the dataset of Figure 1.

A *conditional contingency table*, written

$$\mathbf{ct}(a_{i(1)} \ldots a_{i(n)} \mid a_{j(1)} = u_1, \ldots a_{j(p)} = u_p) \qquad (2)$$

is the contingency table for the subset of records in the dataset that match the query to the right of the | symbol. For example,

$$\mathbf{ct}(a_1, a_3 \mid a_2 = 3) =$$

| $a_1$ | $a_3$ | # |
|---|---|---|
| 1 | 1 | 2 |
| 1 | 2 | 0 |
| 2 | 1 | 2 |
| 2 | 2 | 0 |

Contingency tables are used in a variety of machine learning applications, including building the probability tables for Bayes nets and evaluating candidate conjunctive rules in rule learning algorithms (Quinlan, 1990; Clark & Niblett, 1989). It would thus be desirable to be able to perform such counting efficiently.

If we are prepared to pay a one-time cost for building a caching data structure, then it is easy to suggest a mechanism for doing counting in constant time. For each possible query, we precompute the contingency table. The total amount of numbers stored in memory for such a data structure would be $\Pi_{i=1}^{M}(n_i + 1)$, which even for our humble dataset of Figure 1 is 45, as revealed by Figure 2. For a real dataset with more than ten attributes of medium arity, or fifteen binary attributes, this is far too large to fit in main memory.

We would like to retain the speed of precomputed contingency tables without incurring an intractable memory demand. That is the subject of this paper.

## 2. Cache Reduction 1: The Dense *AD*tree for Caching Sufficient Statistics

First we will describe the *AD*tree, the data structure we will use to represent the set of all possible counts. Our initial simplified description is an obvious tree representation that does not yield any immediate memory savings,[1] but will later provide several opportunities for cutting off zero counts and redundant counts. This structure is shown in Figure 3. An *AD*tree node (shown as a rectangle) has child nodes called "Vary nodes" (shown as ovals). Each *AD*node represents a query and stores the number of records that match the query (in the $C = \#$ field). The Vary $a_j$ child of an *AD*node has one child for each of the $n_j$ values of attribute $a_j$. The $k$th such child represents the same query as Vary $a_j$'s parent, with the additional constraint that $a_j = k$.

---

1. The SE-tree (Rymon, 1993) is a similar data structure.

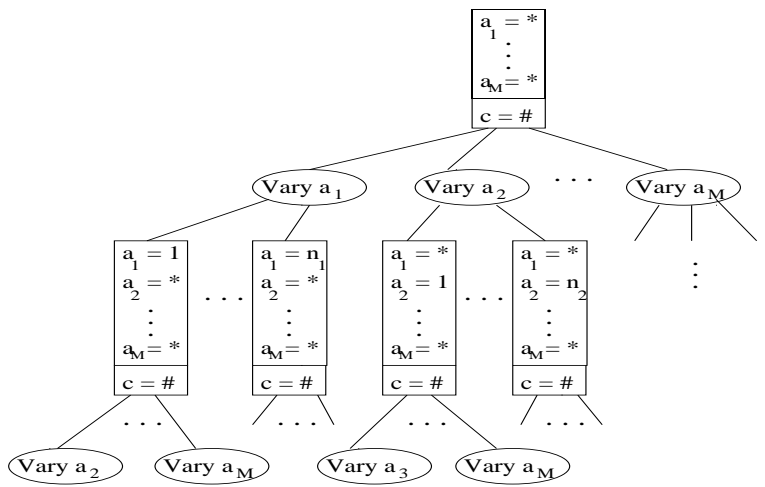

Figure 3: The top *AD*nodes of an *AD*tree, described in the text.

Notes regarding this structure:

- Although drawn on the diagram, the description of the query (e.g., $a_1 = 1, a_2 = *, ..a_M = *$ on the leftmost *AD*node of the second level) is not explicitly recorded in the *AD*node. The contents of an *AD*node are simply a count and a set of pointers to the Vary $a_j$ children.

  The contents of a Vary $a_j$ node are a set of pointers to *AD*nodes.

- The cost of looking up a count is proportional to the number of instantiated variables in the query. For example, to look up $C(a_7 = 2, a_{13} = 1, a_{22} = 3)$ we would follow the following path in the tree: Vary $a_7 \rightarrow a_7 = 2 \rightarrow$ Vary $a_{13} \rightarrow a_{13} = 1 \rightarrow$ Vary $a_{22} \rightarrow a_{22} = 3$. Then the count is obtained from the resulting node.

- Notice that if a node $ADN$ has Vary $a_i$ as its parent, then $ADN$'s children are

  Vary $a_{i+1}$ Vary $a_{i+2}$ ... Vary $a_M$.

  It is not necessary to store Vary nodes with indices below $i+1$ because that information can be obtained from another path in the tree.

## 2.1 Cutting off nodes with counts of zero

As described, the tree is not sparse and will contain exactly $\Pi_{i=1}^{M}(n_i + 1)$ nodes. Sparseness is easily achieved by storing a *NULL* instead of a node for any query that matches zero records. All of the specializations of such a query will also have a count of zero and they will not appear anywhere in the tree. For some datasets this can reduce the number of numbers that need to be stored. For example, the dataset in Figure 1, which previously needed 45 numbers to represent all contingency tables, will now only need 22 numbers.

## 3. Cache Reduction II: The Sparse *AD*tree

It is easy to devise datasets for which there is no benefit in failing to store counts of zero. Suppose we have M binary attributes and $2^M$ records in which the $k$th record is the bits of the binary representation of $k$. Then no query has a count of zero and the tree contains $3^M$ nodes. To reduce the tree size despite this, we will take advantage of the observation that very many of the counts stored in the above tree are redundant.

Each Vary $a_j$ node in the above *AD*tree stores $n_j$ subtrees—one subtree for each value of $a_j$. Instead, we will find the most common of the values of $a_j$ (call it *MCV*) and store a *NULL* in place of the *MCV*th subtree. The remaining $n_j - 1$ subtrees will be represented as before. An example for a simple dataset is given in Figure 4. Each Vary $a_j$ node now records which of its values is most common in a *MCV* field. Appendix B describes the straightforward algorithm for building such an *AD*tree.

As we will see in Section 4, it is still possible to build full exact contingency tables (or give counts for specific queries) in time that is only slightly longer than for the full *AD*tree of Section 2. But first let us examine the memory consequences of this representation.

Appendix A shows that for binary attributes, given $M$ attributes and $R$ records, the number of nodes needed to store the tree is bounded above by $2^M$ in the worst case (and much less if $R < 2^M$). In contrast, the amount of memory needed by the dense tree of Section 2 is $3^M$ in the worst case.

Notice in Figure 4 that the *MCV* value is context dependent. Depending on constraints on parent nodes, $a_2$'s *MCV* is sometimes 1 and sometimes 2. This context dependency can provide dramatic savings if (as is frequently the case) there are correlations among the attributes. This is discussed further in Appendix A.

## 4. Computing Contingency Tables from the Sparse *AD*tree

Given an *AD*tree, we wish to be able to quickly construct contingency tables for any arbitrary set of attributes $\{a_{i(1)} \ldots a_{i(n)}\}$.

Notice that a conditional contingency table $\mathbf{ct}(a_{i(1)} \ldots a_{i(n)} \mid Query)$ can be built recursively. We first build

$$\mathbf{ct}(a_{i(2)} \ldots a_{i(n)} \mid a_{i(1)} = 1, Query)$$
$$\mathbf{ct}(a_{i(2)} \ldots a_{i(n)} \mid a_{i(1)} = 2, Query)$$
$$\vdots$$
$$\mathbf{ct}(a_{i(2)} \ldots a_{i(n)} \mid a_{i(1)} = n_{i(1)}, Query)$$

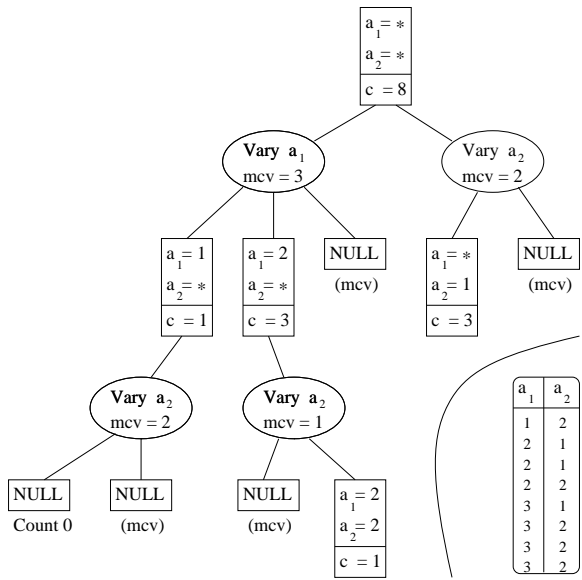

Figure 4: A sparse $AD$tree built for the dataset shown in the bottom right. The most common value for $a_1$ is 3, and so the $a_1 = 3$ subtree of the Vary $a_1$ child of the root node is NULL. At each of the Vary $a_2$ nodes the most common child is also set to NULL (which child is most common depends on the context).

For example, to build $\mathbf{ct}(a_1, a_3)$ using the dataset in Figure 1, we can build $\mathbf{ct}(a_3 \mid a_1 = 1)$ and $\mathbf{ct}(a_3 \mid a_1 = 2)$ and combine them as in Figure 5.

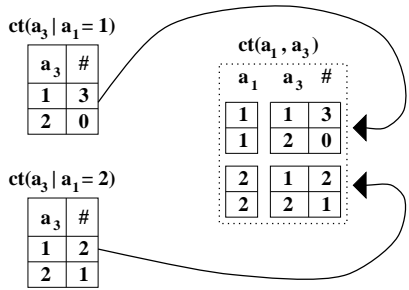

Figure 5: An example (using numbers from Figure 1) of how contingency tables can be combined recursively to form larger contingency tables.

When building a conditional contingency table from an $AD$tree, we will not need to explicitly specify the query condition. Instead, we will supply an $AD$node of the $AD$tree, which implicitly is equivalent information. The algorithm is:

**MakeContab**( { $a_{i(1)} \ldots a_{i(n)}$ } , $ADN$)
      Let $VN$ := The Vary $a_{i(1)}$ subnode of $ADN$.
      Let $MCV$ := $VN.MCV$.
      For $k$ := $1, 2, \ldots, n_{i(1)}$
            If $k \neq MCV$
                  Let $ADN_k$ := The $a_{i(1)} = k$ subnode of $VN$.
                  $CT_k$ := **MakeContab**($\{a_{i(2)} \ldots a_{i(n)}\}, ADN_k$).

      $CT_{\mathrm{MCV}}$ := (calculated as explained below)

      Return the concatenation of $CT_1 \ldots CT_{n_{i(1)}}$.

The base case of this recursion occurs when the first argument is empty, in which case we return a one-element contingency table containing the count associated with the current $AD$node, $ADN$.

There is an omission in the algorithm. In the iteration over $k \in \{1, 2, \ldots n_{i(1)}\}$ we are unable to compute the conditional contingency table for $CT_{\mathrm{MCV}}$ because the $a_{i(1)} = MCV$ subtree is deliberately missing as per Section 3. What can we do instead?

We can take advantage of the following property of contingency tables:

$$\mathbf{ct}(a_{i(2)} \ldots a_{i(n)} \mid Query) = \sum_{k=1}^{n_{i(1)}} \mathbf{ct}(a_{i(2)} \ldots a_{i(n)} \mid a_{i(1)} = k, Query) \qquad (3)$$

The value $\mathbf{ct}(a_{i(2)} \ldots a_{i(n)} \mid Query)$ can be computed from within our algorithm by calling

$$\mathbf{MakeContab}(\{a_{i(2)} \ldots a_{i(n)}\}, ADN) \qquad (4)$$

and so the missing conditional contingency table in the algorithm can be computed by the following row-wise subtraction:

$$CT_{\mathrm{MCV}} := \mathbf{MakeContab}(\{a_{i(2)} \ldots a_{i(n)}\}, ADN) - \sum_{k \neq MCV} CT_k \qquad (5)$$

Frequent Sets (Agrawal et al., 1996), which are traditionally used for learning association rules, can also be used for computing counts. A recent paper (Mannila & Toivonen, 1996), which also employs a similar subtraction trick, calculates counts from Frequent Sets. In Section 8 we will discuss the strengths and weaknesses of Frequent Sets in comparison with $AD$trees.

## 4.1 Complexity of building a contingency table

What is the cost of computing a contingency table? Let us consider the theoretical worst-case cost of computing a contingency table for $n$ attributes each of arity $k$—note that this cost is unrealistically pessimistic (except when $k = 2$), because most contingency tables are sparse, as discussed later. The assumption that all attributes have the same arity, $k$, is made to simplify the calculation of the worst-case cost, but is not needed by the code.

A contingency table for $n$ attributes has $k^n$ entries. Write $C(n) =$ the cost of computing such a contingency table. In the top-level call of **MakeContab** there are $k$ calls to build contingency tables from $n - 1$ attributes: $k - 1$ of these calls are to build $CT(a_{i(2)} \ldots a_{i(n)} \mid a_{i(1)} = j, Query)$ for every $j$ in $\{1, 2, \ldots k\}$ *except* the MCV, and the final call is to build $CT(a_{i(2)} \ldots a_{i(n)} \mid Query)$. Then there will be $k - 1$ subtractions of contingency tables, which will each require $k^{n-1}$ numeric subtractions. So we have

$$C(0) = 1 \qquad (6)$$

$$C(n) = kC(n-1) + (k-1)k^{n-1} \quad \text{if } n > 0 \qquad (7)$$

The solution to this recurrence relation is $C(n) = (1 + n(k-1))k^{n-1}$; this cost is loglinear in the size of the contingency table. By comparison, if we used no cached data structure, but simply counted through the dataset in order to build a contingency table we would need $O(nR + k^n)$ operations where $R$ is the number of records in the dataset. We are thus cheaper than the standard counting method if $k^n \ll R$. We are interested in large datasets in which $R$ may be more than $100,000$. In such a case our method will present a several order of magnitude speedup for, say, a contingency table of eight binary attributes. Notice that this cost is **independent of M**, the total number of attributes in the dataset, and only depends upon the (almost always much smaller) number of attributes $n$ requested for the contingency table.

## 4.2 Sparse representation of contingency tables

In practice, we do not represent contingency tables as multidimensional arrays, but rather as tree structures. This gives both the slow counting approach and the $AD$tree approach a substantial computational advantage in cases where the contingency table is sparse, i.e. has many zero entries. Figure 6 shows such a sparse contingency table representation. This can mean average-case behavior is much faster than worst case for contingency tables with large numbers of attributes or high-arity attributes.

Indeed, our experiments in Section 7 show costs rising much more slowly than $O(nk^{n-1})$ as $n$ increases. Note too that when using a sparse representation, the worst-case for **MakeContab** is now $O(\min(nR, nk^{n-1}))$ because $R$ is the maximum possible number of non-zero contingency table entries.

## 5. Cache Reduction III: Leaf-Lists

We now introduce a scheme for further reducing memory use. It is not worth building the $AD$tree data structure for a small number of records. For example, suppose we have 15 records and 40 binary attributes. Then the analysis in Appendix A shows us that in the worst case the $AD$tree might require 10701 nodes. But computing contingency tables using the resulting $AD$tree would, with so few records, be no faster than the conventional counting approach, which would merely require us to retain the dataset in memory.

Aside from concluding that $AD$trees are not useful for very small datasets, this also leads to a final method for saving memory in large $AD$trees. Any $AD$tree node with fewer than $R_{\min}$ records does not expand its subtree. Instead it maintains a list of pointers into the original dataset, explicitly listing those records that match the current $AD$node. Such a list of pointers is called a *leaf-list*. Figure 7 gives an example.

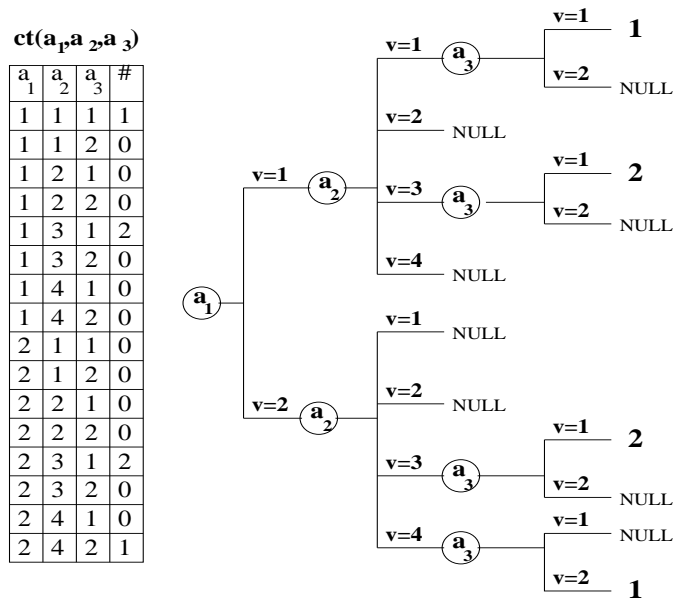

Figure 6: The right hand figure is the sparse representation of the contingency table on the left.

The use of leaf-lists has one minor and two major consequences. The minor consequence is the need to include a straightforward change in the contingency table generating algorithm to handle leaf-list nodes. This minor alteration is not described here. The first major consequence is that now the dataset itself must be retained in main memory so that algorithms that inspect leaf-lists can access the rows of data pointed to in those leaf-lists. The second major consequence is that the $AD$tree may require much less memory. This is documented in Section 7 and worst-case bounds are provided in Appendix A.

## 6. Using $AD$trees for Machine Learning

As we will see in Section 7, the $AD$tree structure can substantially speed up the computation of contingency tables for large real datasets. How can machine learning and statistical algorithms take advantage of this? Here we provide three examples: Feature Selection, Bayes net scoring and rule learning. But it seems likely that many other algorithms can also benefit, for example stepwise logistic regression, GMDH (Madala & Ivakhnenko, 1994), and text classification. Even decision tree (Quinlan, 1983; Breiman, Friedman, Olshen, & Stone, 1984) learning may benefit. [2] In future work we will also examine ways to speed up nearest neighbor and other memory-based queries using $AD$trees.

---

2. This depends on whether the cost of initially building the $AD$tree can be amortized over many runs of the decision tree algorithm. Repeated runs of decision tree building can occur if one is using the wrapper model of feature selection (John, Kohavi, & Pfleger, 1994), or if one is using a more intensive search over tree structures than the traditional greedy search (Quinlan, 1983; Breiman et al., 1984)

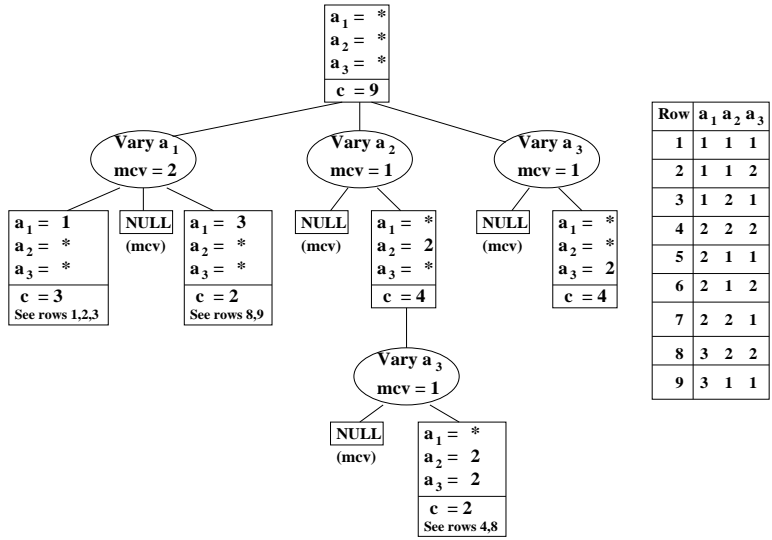

Figure 7: An *AD*tree built using leaf-lists with $R_{\min} = 4$. Any node matching 3 or fewer records is not expanded, but simply records a set of pointers into the dataset (shown on the right).

## 6.1 Datasets

The experiments used the datasets in Table 1. Each dataset was supplied to us with all continuous attributes already discretized into ranges.

## 6.2 Using *AD*trees for Feature Selection

Given $M$ attributes, of which one is an output that we wish to predict, it is often interesting to ask "which subset of $n$ attributes, $(n < M)$, is the best predictor of the output on the same distribution of datapoints that are reflected in this dataset?" (Kohavi, 1995). There are many ways of scoring a set of features, but a particularly simple one is *information gain* (Cover & Thomas, 1991).

Let $a_{out}$ be the attribute we wish to predict and let $a_{i(1)} \ldots a_{i(n)}$ be the set of attributes used as inputs. Let $X$ be the set of possible assignments of values to $a_{i(1)} \ldots a_{i(n)}$ and write $Assign_k \in X$ as the $k$th such assignment. Then

$$InfoGain = \sum_{v=1}^{n_{out}} f\left(\frac{C(a_{out} = v)}{R}\right) - \sum_{k=1}^{|X|} \frac{C(Assign_k)}{R} \sum_{v=1}^{n_{out}} f\left(\frac{C(a_{out} = v, Assign_k)}{C(Assign_k)}\right) \quad (8)$$

where $R$ is the number of records in the entire dataset and

$$f(x) = -x \log_2 x \quad (9)$$

The counts needed in the above computation can be read directly from $\mathbf{ct}(a_{out}, a_{i(1)} \ldots a_{i(n)})$.

Searching for the best subset of attributes is simply a question of search among all attribute-sets of size $n$ ($n$ specified by the user). This is a simple example designed to test our counting methods: any practical feature selector would need to penalize the number of rows in the contingency table (else high arity attributes would tend to win).

| Name | $R$ = Num. Records | $M$ = Num. Attributes | |
|------|--------|----------|--|
| ADULT1 | 15,060 | 15 | The small "Adult Income" dataset placed in the UCI repository by Ron Kohavi (Kohavi, 1996). Contains census data related to job, wealth, and nationality. Attribute arities range from 2 to 41. In the UCI repository this is called the Test Set. Rows with missing values were removed. |
| ADULT2 | 30,162 | 15 | The same kinds of records as above but with different data. The Training Set. |
| ADULT3 | 45,222 | 15 | ADULT1 and ADULT2 concatenated. |
| CENSUS1 | 142,421 | 13 | A larger dataset based on a different census, also provided by Ron Kohavi. |
| CENSUS2 | 142,421 | 15 | The same data as CENSUS1, but with the addition of two extra, high-arity attributes. |
| BIRTH | 9,672 | 97 | Records concerning a very wide number of readings and factors recorded at various stages during pregnancy. Most attributes are binary, and 70 of the attributes are very sparse, with over 95% of the values being FALSE. |
| SYNTH | 30K–500K | 24 | Synthetic datasets of entirely binary attributes generated using the Bayes net in Figure 8. |

Table 1: Datasets used in experiments.

### 6.3 Using $AD$trees for Bayes Net Structure Discovery

There are many possible Bayes net learning tasks, all of which entail counting, and hence might be speeded up by $AD$trees. In this paper we present experimental results for the particular example of scoring the structure of a Bayes net to decide how well it matches the data.

We will use maximum likelihood scoring with a penalty for the number of parameters. We first compute the probability table associated with each node. Write $Parents(j)$ for the parent attributes of node $j$ and write $X_j$ as the set of possible assignments of values to $Parents(j)$. The maximum likelihood estimate for

$$P(a_j = v \mid X_j) \tag{10}$$

is estimated as

$$\frac{C(a_j = v, X_j)}{C(X_j)} \tag{11}$$

and all such estimates for node $j$'s probability tables can be read from $\mathbf{ct}(a_j, Parents(j))$.

The next step in scoring a structure is to decide the likelihood of the data given the probability tables we computed and to penalize the number of parameters in our network (without the penalty the likelihood would increase every time a link was added to the

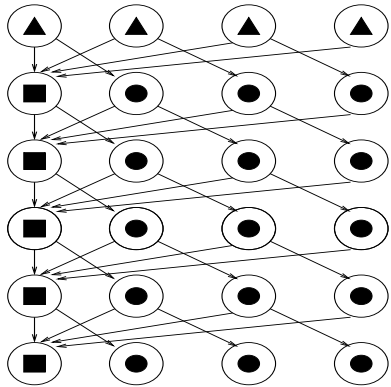

Figure 8: A Bayes net that generated our SYNTH datasets. There are three kinds of nodes. The nodes marked with triangles are generated with $P(a_i = 1) = 0.8, P(a_i = 2) = 0.2$. The square nodes are deterministic. A square node takes value 2 if the sum of its four parents is even, else it takes value 1. The circle nodes are probabilistic functions of their single parent, defined by $P(a_i = 2 \mid Parent = 1) = 0$ and $P(a_i = 2 \mid Parent = 2) = 0.4$. This provides a dataset with fairly sparse values and with many interdependencies.

network). The penalized log-likelihood score (Friedman & Yakhini, 1996) is

$$- N_{\text{params}} \log(R)/2 + R \sum_{j=1}^{M} \sum_{Asgn \in X_j} \sum_{v=1}^{n_j} P(a_j = v \wedge Asgn) \log P(a_j = v \mid Asgn) \qquad (12)$$

where $N_{\text{params}}$ is the total number of probability table entries in the network.

We search among structures to find the best score. In these experiments we use random-restart stochastic hill climbing in which the operations are random addition or removal of a network link or randomly swapping a pair of nodes. The latter operation is necessary to allow the search algorithm to choose the best ordering of nodes in the Bayes net. Stochastic searches such as this are a popular method for finding Bayes net structures (Friedman & Yakhini, 1996). Only the probability tables of the affected nodes are recomputed on each step.

Figure 9 shows the Bayes net structure returned by our Bayes net structure finder after 30,000 iterations of hill climbing.

## 6.4 Using $AD$trees for Rule Finding

Given an output attribute $a_{out}$ and a distinguished value $v_{\text{out}}$, rule finders search among conjunctive queries of the form

$$Assign = (a_{i(1)} = v_1 \dots a_{i(n)} = v_n) \qquad (13)$$

```
attribute      score      np
relationship   2.13834      2   pars = <no parents>
class          0.643388    36   pars = relationship
sex            0.511666    24   pars = relationship class
capital-gain   0.0357936   30   pars = class
hours-per-week 0.851964    48   pars = relationship class sex
marital-status 0.762479    72   pars = relationship sex
education-num  1.0941      26   pars = class
capital-loss   0.22767     10   pars = class
age            0.788001    28   pars = marital-status
race           0.740212    18   pars = relationship education-num
education      1.71784     36   pars = relationship education-num
workclass      1.33278    108   pars = relationship hours-per-week education-num
native-country 0.647258    30   pars = education-num race
fnlwgt         0.0410872   40   pars = <no parents>
occupation     2.66097    448   pars = class sex education workclass

Score is 435219
The search took 226 seconds.
```

Figure 9: Output from the Bayes structure finder running on the ADULT2 dataset. *Score* is the contribution to the sum in Equation 12 due to the specified attribute. *np* is the number of entries in the probability table for the specified attribute.

to find the query that maximizes the estimated value

$$P(a_{out} = v_{\text{out}} \mid Assign) = \frac{C(a_{out} = v_{\text{out}}, Assign)}{C(Assign)} \qquad (14)$$

To avoid rules without significant support, we also insist that $C(Assign)$ (the number of records matching the query) must be above some threshold $S_{\min}$.

In these experiments we implement a brute force search that looks through all possible queries that involve a user-specified number of attributes, $n$. We build each $\mathbf{ct}(a_{out}, a_{i(1)} \dots a_{i(n)})$ in turn (there are $M$ *choose* $n$ such tables), and then look through the rows of each table for all queries using the $a_{i(1)} \dots a_{i(n)}$ that have greater than minimum support $S_{\min}$. We return a priority queue of the highest scoring rules. For instance on the ADULT2 dataset, the best rule for predicting "class" from 4 attributes was:

score = 0.965 (218/226), workclass = Private, education-num = above12, marital-status = Married-civ-spouse, capital-loss = above1600 $\Rightarrow$ class $\geq$ 50k

## 7. Experimental Results

Let us first examine the memory required by an *AD*tree on our datasets. Table 2 shows us, for example, that the ADULT2 dataset produced an *AD*tree with 95,000 nodes. The tree required almost 11 megabytes of memory. Among the three ADULT datasets, the size of the tree varied approximately linearly with the number of records.

Unless otherwise specified, in all the experiments in this section, the ADULT datasets used no leaf-lists. The BIRTH and SYNTHETIC datasets used leaf-lists of size $R_{\min} = 16$ by default. The BIRTH dataset, with its large number of sparse attributes, required a modest 8 megabytes to store the tree—many magnitudes below the worst-case bounds. Among the synthetic datasets, the tree size increased sublinearly with the dataset size. This indicates that as the dataset gets larger, novel records (which may cause new nodes to appear in the tree) become less frequent.

| Dataset | M | R | Nodes | Megabytes | Build Time |
|---------|----|--------|--------|-----------|------------|
| CENSUS1 | 13 | 142521 | 24007 | 1.5 | 17 |
| CENSUS2 | 15 | 142521 | 209577 | 13.2 | 32 |
| ADULT1 | 15 | 15060 | 58200 | 7.0 | 6 |
| ADULT2 | 15 | 30162 | 94900 | 10.9 | 10 |
| ADULT3 | 15 | 45222 | 162900 | 15.5 | 15 |
| BIRTH | 97 | 9672 | 87400 | 7.9 | 14 |
| SYN30K | 24 | 30000 | 34100 | 2.8 | 8 |
| SYN60K | 24 | 60000 | 59300 | 4.9 | 17 |
| SYN125K | 24 | 125000 | 95400 | 7.9 | 36 |
| SYN250K | 24 | 250000 | 150000 | 12.4 | 73 |
| SYN500K | 24 | 500000 | 219000 | 18.2 | 150 |

Table 2: The size of *AD*trees for various datasets. $M$ is the number of attributes. $R$ is the number of records. *Nodes* is the number of nodes in the *AD*tree. *Megabytes* is the amount of memory needed to store the tree. *Build Time* is the number of seconds needed to build the tree (to the nearest second).

Table 3 shows the costs of performing 30,000 iterations of Bayes net structure searching. All experiments were performed on a 200Mhz Pentium Pro machine with 192 megabytes of main memory. Recall that each Bayes net iteration involves one random change to the network and so requires recomputation of one contingency table (the exception is the first iteration, in which all nodes must be computed). This means that the time to run 30,000 iterations is essentially the time to compute 30,000 contingency tables. Among the ADULT datasets, the advantage of the *AD*tree over conventional counting ranges between a factor of 19 to 32. Unsurprisingly, the computational costs for ADULT increase sublinearly with dataset size for the *AD*tree but linearly for the conventional counting. The computational advantages and the sublinear behavior are much more pronounced for the synthetic data.

Next, Table 4 examines the effect of leaf-lists on the ADULT2 and BIRTH datasets. For the ADULT dataset, the byte size of the tree decreases by a factor of 5 when leaf-lists are increased from 1 to 64. But the computational cost of running the Bayes search increases by only 25%, indicating a worth-while tradeoff if memory is scarce.

The Bayes net scoring results involved the average cost of computing contingency tables of many different sizes. The following results in Tables 5 and 6 make the savings for fixed size attribute sets easier to discern. These tables give results for the feature selection and rule finding algorithms, respectively. The biggest savings come from small attribute sets. Computational savings for sets of size one or two are, however, not particularly interesting since all such counts could be cached by straightforward methods without needing any tricks. In all cases, however, we do see large savings, especially for the BIRTH data. Datasets with larger numbers of rows would, of course, reveal larger savings.

| Dataset | M | R | *AD*tree Time | Regular Time | Speedup Factor |
|---------|---|---|---------------|--------------|----------------|
| CENSUS1 | 13 | 142521 | 162 | 48300 | 298.1 |
| CENSUS2 | 15 | 142521 | 783 | 27000 | 34.5 |
| ADULT1 | 15 | 15060 | 150 | 2850 | 19 |
| ADULT2 | 15 | 30162 | 226 | 5160 | 22.8 |
| ADULT3 | 15 | 45222 | 220 | 7140 | 32.5 |
| BIRTH | 97 | 9672 | 23 | 2820 | 122.6 |
| SYN30K | 24 | 30000 | 32 | 10410 | 325.3 |
| SYN60K | 24 | 60000 | 32 | 18360 | 573.8 |
| SYN125K | 24 | 125000 | 34 | 42840 | 1260.0 |
| SYN250K | 24 | 250000 | 35 | 88830 | 2538.0 |
| SYN500K | 24 | 500000 | 36 | 155158 | 4298.3 |

Table 3: The time (in seconds) to perform 30,000 hill-climbing iterations searching for the best Bayes net structure. *ADtree Time* is the time when using the *AD*tree and *Regular Time* is the time taken when using the conventional probability table scoring method of counting through the dataset. *Speedup Factor* is the number of times by which the *AD*tree method is faster than the conventional method. The *AD*tree times do not include the time for building the *AD*tree in the first place (given in Table 2). A typical use of *AD*trees will build the tree only once and then be able to use it for many data analysis operations, and so its building cost can be amortized. In any case, even including tree building cost would have only a minor impact on the results.

| $R_{min}$ | ADULT2 | | | | BIRTH | | | |
|-----------|--------|---------|-------|--------|-------|---------|-------|--------|
| | #Mb | #nodes | Build Secs | Search Secs | #Mb | #nodes | Build Secs | Search Secs |
| 1 | 10.87 | 94,872 | 13 | 225 | | | | |
| 2 | 8.46 | 86,680 | 11 | 223 | | | | |
| 4 | 6.30 | 75,011 | 9 | 223 | 27.60 | 245,722 | 32 | 25 |
| 8 | 4.62 | 62,095 | 7 | 224 | 14.90 | 152,409 | 21 | 22 |
| 16 | 3.37 | 49,000 | 6 | 232 | 7.95 | 87,415 | 15 | 23 |
| 32 | 2.55 | 37,790 | 5 | 245 | 4.30 | 46,777 | 11 | 26 |
| 64 | 1.98 | 27,726 | 5 | 274 | 2.42 | 23,474 | 8 | 31 |
| 128 | 1.59 | 18,903 | 3 | 331 | 1.48 | 11,554 | 7 | 38 |
| 256 | 1.38 | 12,539 | 3 | 420 | 0.95 | 5,150 | 5 | 61 |
| 512 | 1.21 | 7,336 | 3 | 586 | 0.65 | 2,237 | 4 | 100 |
| 1024 | 1.05 | 3,928 | 3 | 881 | 0.45 | 887 | 3 | 170 |
| 2048 | 0.95 | 1,890 | 1 | 1231 | 0.28 | 246 | 1 | 270 |
| 4096 | 0.86 | 780 | 1 | 1827 | 0.28 | 201 | 2 | 303 |

Table 4: Investigating the effect of the $R_{min}$ parameter on the ADULT2 dataset and the BIRTH dataset. *#Mb* is the memory used by the *AD*tree. *#nodes* is the number of nodes in the *AD*tree. *Build Secs* is the time to build the *AD*tree. *Search Secs* is the time needed to perform 30,000 iterations of the Bayes net structure search.

| | ADULT2 | | | | BIRTH | | | |
|---|---|---|---|---|---|---|---|---|
| Number Attributes | Number Attribute Sets | *AD*tree Time | Regular Time | Speedup Factor | Number Attribute Sets | *AD*tree Time | Regular Time | Speedup Factor |
| 1 | 14 | .000071 | .048 | 675.0 | 96 | .000018 | .015 | 841 |
| 2 | 91 | .00054 | .067 | 124.0 | 4,560 | .000042 | .021 | 509 |
| 3 | 364 | .0025 | .088 | 34.9 | 142,880 | .000093 | .028 | 298 |
| 4 | 1,001 | .0083 | .11 | 13.4 | 3,321,960 | .00019 | | |
| 5 | 2,002 | .023 | .14 | 6.0 | 61,124,064 | .00033 | | |

Table 5: The time taken to search among all attribute sets of a given size (Number Attributes) for the set that gives the best information gain in predicting the output attribute. The times, in seconds, are the average evaluation times per attribute-set.

| | ADULT2 | | | | BIRTH | | | |
|---|---|---|---|---|---|---|---|---|
| Number Attributes | Number Rules | *AD*tree Time | Regular Time | Speedup Factor | Number Rules | *AD*tree Time | Regular Time | Speedup Factor |
| 1 | 116 | .000019 | .0056 | 295.0 | 194 | .000025 | .0072 | 286 |
| 2 | 4,251 | .000019 | .0014 | 75.3 | 17,738 | .000021 | .0055 | 259 |
| 3 | 56,775 | .000024 | .00058 | 23.8 | 987,134 | .000022 | .0040 | 186 |
| 4 | 378,984 | .000031 | .00030 | 9.8 | 37,824,734 | .000024 | .0030 | 127 |
| 5 | 1,505,763 | .000042 | .00019 | 4.7 | 1,077,672,055 | .000026 | | |

Table 6: The time taken to search among all rules of a given size (Number Attributes) for the highest scoring rules for predicting the output attribute. The times, in seconds, are the average evaluation time per rule.

## 8. Alternative Data Structures

### 8.1 Why not use a $k$d-tree?

$k$d-trees can be used for accelerating learning algorithms (Omohundro, 1987; Moore et al., 1997). The primary difference is that a $k$d-tree node splits on only one attribute instead of all attributes. This results in much less memory (linear in the number of records). But counting can be expensive. Suppose, for example, that level one of the tree splits on $a_1$, level two splits on $a_2$, etc. Then, in the case of binary variables, if we have a query involving only attributes $a_{20}$ and higher, we have to explore all paths in the tree down to level 20. With datasets of fewer than $2^{20}$ records this may be no cheaper than performing a linear search through the records. Another possibility, $R$-trees (Guttman, 1984; Roussopoulos & Leifker, 1985), store databases of $M$-dimensional geometric objects. However, in this context, they offer no advantages over $k$d-trees.

### 8.2 Why not use a Frequent Set finder?

Frequent Set finders (Agrawal et al., 1996) are typically used with very large databases of millions of records containing very sparse binary attributes. Efficient algorithms exist for finding all subsets of attributes that co-occur with value TRUE in more than a fixed number (chosen by the user, and called the *support*) of records. Recent research (Mannila & Toivonen, 1996) suggests that such Frequent Sets can be used to perform efficient counting. In the case where *support* $= 1$, all such Frequent Sets are gathered and, if counts of each Frequent Set are retained, this is equivalent to producing an $AD$tree in which instead of performing a node cutoff for the most common value, the cutoff always occurs for value FALSE.

The use of Frequent Sets in this way would thus be very similar to the use of $AD$trees, with one advantage and one disadvantage. The advantage is that efficient algorithms have been developed for building Frequent Sets from a small number of sequential passes through data. The $AD$tree requires random access to the dataset while it is being built, and for its leaf-lists. This is impractical if the dataset is too large to reside in main memory and is accessed through database queries.

The disadvantage of Frequent Sets in comparison with $AD$trees is that, under some circumstances, the former may require much more memory. Assume the value 2 is rarer than 1 throughout all attributes in the dataset and assume reasonably that we thus choose to find all Frequent Sets of 2s. Unnecessarily many sets will be produced if there are correlations. In the extreme case, imagine a dataset in which 30% of the values are 2, 70% are 1 and attributes perfectly correlated—all values in each record are identical. Then, with $M$ attributes there would be $2^M$ Frequent Sets of 2s. In contrast, the $AD$tree would only contain $M + 1$ nodes. This is an extreme example, but datasets with much weaker inter-attribute correlations can similarly benefit from using an $AD$tree.

Leaf-lists are another technique to reduce the size of $AD$trees further. They could also be used for the Frequent Set representation.

### 8.3 Why not use hash tables?

If we knew that only a small set of contingency tables would ever be requested, instead of all possible contingency tables, then an $AD$tree would be unnecessary. It would be better to remember this small set of contingency tables explicitly. Then, some kind of tree structure could be used to index the contingency tables. But a hash table would be equally time efficient and require less space. A hash table coding of individual counts in the contingency tables would similarly allow us to use space proportional only to the number of non-zero entries in the stored tables. But for representing sufficient statistics to permit fast solution to any contingency table request, the $AD$tree structure remains more memory efficient than the hash-table approach (or any method that stores all non-zero counts) because of the memory reductions when we exploit the ignoring of most common values.

## 9. Discussion

### 9.1 What about numeric attributes?

The $AD$tree representation is designed entirely for symbolic attributes. When faced with numeric attributes, the simplest solution is to discretize them into a fixed finite set of values which are then treated as symbols, but this is of little help if the user requests counts for queries involving inequalities on numeric attributes. In future work we will evaluate the use of structures combining elements from multiresolution $k$d-trees of real attributes (Moore et al., 1997) with $AD$trees.

### 9.2 Algorithm-specific counting tricks

Many algorithms that count using the conventional "linear" method have algorithm-specific ways of accelerating their performance. For example, a Bayes net structure finder may try to remember all the contingency tables it has tried previously in case it needs to re-evaluate them. When it deletes a link, it can deduce the new contingency table from the old one without needing a linear count.

In such cases, the most appropriate use of the $AD$tree may be as a lazy caching mechanism. At birth, the $AD$tree consists only of the root node. Whenever the structure finder needs a contingency table that cannot be deduced from the current $AD$tree structure, the appropriate nodes of the $AD$tree are expanded. The $AD$tree then takes on the role of the algorithm-specific caching methods, while (in general) using up much less memory than if all contingency tables were remembered.

### 9.3 Hard to update incrementally

Although the tree can be built cheaply (see the experimental results in Section 7), and although it can be built lazily, the $AD$tree cannot be updated cheaply with a new record. This is because one new record may match up to $2^M$ nodes in the tree in the worst case.

### 9.4 Scaling up

The $AD$tree representation can be useful for datasets of the rough size and shape used in this paper. On the first datasets we have looked at—the ones described in this paper—we have shown empirically that the sizes of the $AD$trees are tractable given real noisy data. This included one dataset with 97 attributes. It is the extent to which the attributes are skewed in their values and correlated with each other that enables the $AD$tree to avoid approaching its worse-case bounds. The main technical contribution of this paper is the trick that allows us to prune off most-common-values. Without it, skewedness and correlation would hardly help at all. [3] The empirical contribution of this paper has been to show that the actual sizes of the $AD$trees produced from real data are vastly smaller than the sizes we would get from the worst-case bounds in Appendix A.

But despite these savings, $AD$trees cannot yet represent all the sufficient statistics for huge datasets with many hundreds of non-sparse and poorly correlated attributes. What should we do if our dataset or our $AD$tree cannot fit into main memory? In the latter case, we could simply increase the size of leaf-lists, trading off decreased memory against increased time to build contingency tables. But if that is inadequate at least three possibilities remain. First, we could build approximate $AD$trees that do not store any information for nodes that match fewer than a threshold number of records. Then approximate contingency tables (complete with error bounds) can be produced (Mannila & Toivonen, 1996). A second possibility is to exploit secondary storage and store deep, rarely visited nodes of the $AD$tree on disk. This would doubtless best be achieved by integrating the machine learning algorithms with current database management tools—a topic of considerable interest in the data mining community (Fayyad et al., 1997). A third possibility, which restricts the size of contingency tables we may ask for, is to refuse to store counts for queries with more than some threshold number of attributes.

## 9.5 What about the cost of building the tree?

In practice, $AD$trees could be used in two ways:

- **One-off.** When a traditional algorithm is required we build the $AD$tree, run the fast version of the algorithm, discard the $AD$tree, and return the results.

- **Amortized.** When a new dataset becomes available, a new $AD$tree is built for it. The tree is then shipped and re-used by anyone who wishes to do real-time counting queries, multivariate graphs and charts, or any machine learning algorithms on any subset of the attributes. The cost of the initial tree building is then amortized over all the times it is used. In database terminology, the process is known as materializing (Harinarayan et al., 1996) and has been suggested as desirable for datamining by several researchers (John & Lent, 1997; Mannila & Toivonen, 1996).

The one-off option is only useful if the cost of building the $AD$tree plus the cost of running the $AD$tree-based algorithm is less than the cost of the original counting-based algorithm. For the intensive machine learning methods studied here, this condition is safely satisfied. But what if we decided to use a less intensive, greedier Bayes net structure finder? Table 7 shows that if we only run for 300 iterations instead of 30,000[4] and if we account for a one-off $AD$tree building cost, then the relative speedup of using $AD$trees declines greatly.

---

3. Without pruning, on all of our datasets we ran out of memory on a 192 megabyte machine before we had built even 1% of the tree, and it is easy to show that the BIRTH dataset would have needed to store more than $10^{30}$ nodes.

| Dataset | Speedup ignoring build-time, 30,000 iterations | Speedup allowing for build-time, 30,000 iterations | Speedup allowing for build-time, 300 iterations |
|---|---|---|---|
| CENSUS1 | 298.15 | 269.83 | 25.94 |
| CENSUS2 | 34.48 | 33.13 | 6.78 |
| ADULT1 | 19.00 | 18.27 | 3.80 |
| ADULT2 | 22.83 | 21.86 | 4.21 |
| ADULT3 | 32.45 | 30.38 | 4.15 |
| BIRTH | 122.61 | 76.22 | 1.98 |
| SYN30K | 325.31 | 260.25 | 12.51 |
| SYN60K | 573.75 | 374.69 | 10.60 |
| SYN125K | 1260.00 | 612.00 | 11.79 |
| SYN250K | 2538.00 | 822.50 | 12.11 |
| SYN500K | 4309.94 | 834.18 | 10.32 |

Table 7: Computational economics of building $AD$trees and using them to search for Bayes net structures using the experiments of Section 7.

To conclude: if the data analysis is intense then there is benefit to using $AD$trees even if they are used in a one-off fashion. If the $AD$tree is used for multiple purposes then its build-time is amortized and the resulting relative efficiency gains over traditional counting are the same for both exhaustive searches and non-exhaustive searches. Algorithms that use non-exhaustive searches include hill-climbing Bayes net learners, greedy rule learners such as CN2 (Clark & Niblett, 1989) and decision tree learners (Quinlan, 1983; Breiman et al., 1984).

## Acknowledgements

This work was sponsored by a National Science Foundation Career Award to Andrew Moore. The authors thank Justin Boyan, Scott Davies, Nir Friedman, and Jeff Schneider for their suggestions, and Ron Kohavi for providing the census datasets.

## Appendix A: Memory Costs

In this appendix we examine the size of the tree. For simplicity, we restrict attention to the case of binary attributes.

### The worst-case number of nodes in an $AD$tree

Given a dataset with $M$ attributes and $R$ records, the worst-case for the $AD$tree will occur if all $2^M$ possible records exist in the dataset. Then, for every subset of attributes there exists exactly one node in the $AD$tree. For example consider the attribute set $\{a_{i(1)} \dots a_{i(n)}\}$, where $i(1) < i(2) < \dots < i(n)$. Suppose there is a node in the tree corresponding to the query $\{a_{i(1)} = v_1 \ldots a_{i(n)} = v_n\}$ for some values $v_1 \ldots v_n$. From the definition of an $AD$tree, and remembering we are only considering the case of binary attributes, we can state:

- $v_1$ is the least common value of $a_{i(1)}$.

- $v_2$ is the least common value of $a_{i(2)}$ among those records that match $(a_{i(1)} = v_1)$.

---

4. Unsurprisingly, the resulting Bayes nets have a highly inferior structure.

$\vdots$

- $v_{k+1}$ is the least common value of $a_{i(k+1)}$ among those records that match $(a_{i(1)} = v_1, \ldots, a_{i(k)} = v_k)$.

So there is at most one such node. Moreover, since our worst-case assumption is that all possible records exist in the database, we see that the $AD$tree will indeed contain this node. Thus, the worst-case number of nodes is the same as the number of possible subsets of attributes: $2^M$.

**The worst-case number of nodes in an $AD$tree with a reasonable number of rows**
It is frequently the case that a dataset has $R \ll 2^M$. With fewer records, there is a much lower worst-case bound on the $AD$tree size. A node at the $k$th level of the tree corresponds to a query involving $k$ attributes (counting the root node as level 0). Such a node can match at most $R2^{-k}$ records because each of the node's ancestors up the tree has pruned off at least half the records by choosing to expand only the least common value of the attribute introduced by that ancestor. Thus, there can be no tree nodes at level $\lfloor \log_2 R \rfloor + 1$ of the tree, because such nodes would have to match fewer than $R2^{-\lfloor \log_2 R \rfloor - 1} < 1$ records. They would thus match no records, making them $NULL$.

The nodes in an $AD$tree must all exist at level $\lfloor \log_2 R \rfloor$ or higher. The number of nodes at level $k$ is at most $\binom{M}{k}$, because every node at level $k$ involves an attribute set of size $k$ and because (given binary attributes) for every attribute set there is at most one node in the $AD$tree. Thus the total number of nodes in the tree, summing over the levels is less than

$$\sum_{k=0}^{\lfloor \log_2 R \rfloor} \binom{M}{k} \text{ bounded above by } O(M^{\lfloor \log_2 R \rfloor} / (\lfloor \log_2 R \rfloor - 1)!) \tag{15}$$

**The number of nodes if we assume skewed independent attribute values**
Imagine that all values of all attributes in the dataset are independent random binary variables, taking value 2 with probability $p$ and taking value 1 with probability $1 - p$. Then the further $p$ is from 0.5, the smaller we can expect the $AD$tree to be. This is because, on average, the less common value of a Vary node will match fraction $\min(p, 1 - p)$ of its parent's records. And, on average, the number of records matched at the $k$th level of the tree will be $R(\min(p, 1 - p))^k$. Thus, the maximum level in the tree at which we may find a node matching one or more records is approximately $\lfloor (\log_2 R) / (-\log_2 q) \rfloor$, where $q = \min(p, 1 - p)$. And so the total number of nodes in the tree is approximately

$$\sum_{k=0}^{\lfloor (\log_2 R) / (-\log_2 q) \rfloor} \binom{M}{k} \text{ bounded above by } O(M^{\lfloor (\log_2 R) / (-\log_2 q) \rfloor} / (\lfloor (\log_2 R) / (-\log_2 q) \rfloor - 1)!) \tag{16}$$

Since the exponent is reduced by a factor of $\log_2(1/q)$, skewedness among the attributes thus brings enormous savings in memory.

**The number of nodes if we assume correlated attribute values**

The $AD$tree benefits from correlations among attributes in much the same way that it benefits from skewedness. For example, suppose that each record was generated by the simple Bayes net in Figure 10, where the random variable $B$ is hidden (not included in the record). Then for $i \neq j$, $P(a_i \neq a_j) = 2p(1-p)$. If $ADN$ is any node in the resulting $AD$tree then the number of records matching any other node two levels below $ADN$ in the tree will be fraction $2p(1-p)$ of the number of records matching $ADN$. From this we can see that the number of nodes in the tree is approximately

$$\sum_{k=0}^{\lfloor (\log_2 R)/(-\log_2 q) \rfloor} \binom{M}{k} \text{ bounded above by } O(M^{\lfloor (\log_2 R)/(-\log_2 q) \rfloor}/(\lfloor (\log_2 R)/(-\log_2 q) \rfloor - 1)!) \tag{17}$$

where $q = \sqrt{2p(1-p)}$. Correlation among the attributes can thus also bring enormous savings in memory even if (as is the case in our example) the marginal distribution of individual attributes is uniform.

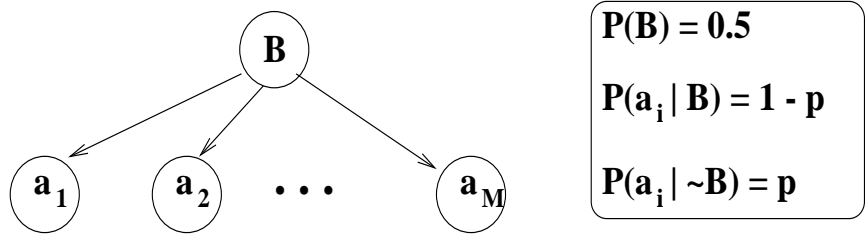

Figure 10: A Bayes net that generates correlated boolean attributes $a_1, a_2 \ldots a_M$.

**The number of nodes for the dense $AD$tree of Section 2**

The dense $AD$trees do not cut off the tree for the most common value of a Vary node. The worst case $AD$tree will occur if all $2^M$ possible records exist in the dataset. Then the dense $AD$tree will require $3^M$ nodes because every possible query (with each attribute taking values 1, 2 or $*$) will have a count in the tree. The number of nodes at the $k$th level of the dense $AD$tree can be $2^k \binom{M}{k}$ in the worst case.

**The number of nodes when using Leaf-lists**

Leaf-lists were described in Section 5. If a tree is built using maximum leaf-list size of $R_{\min}$, then any node in the $AD$tree matching fewer than $R_{\min}$ records is a leaf node. This means that Formulae 15, 16 and 17 can be re-used, replacing $R$ with $R/R_{\min}$. It is important to remember, however, that the leaf nodes must now contain room for $R_{\min}$ numbers instead of a single count.

**Appendix B: Building the $AD$tree**

We define the function **MakeADTree**($a_i$, $RecordNums$) where $RecordNums$ is a subset of $\{1, 2, \ldots, R\}$ ($R$ is the total number of records in the dataset) and where $1 \leq i \leq M$. This makes an $AD$tree from the rows specified in $RecordNums$ in which all $AD$nodes represent queries in which only attributes $a_i$ and higher are used.

> **MakeADTree**($a_i$, $RecordNums$)
>> Make a new $AD$node called $ADN$.
>> $ADN$.COUNT := $| RecordNums |$.
>> For $j$ := $i, i+1, \ldots, M$
>>> $j$th Vary node of $ADN$ := **MakeVaryNode**($a_j, RecordNums$).

**MakeADTree** uses the function **MakeVaryNode**, which we now define:

> **MakeVaryNode**($a_i$, $RecordNums$)
>> Make a new Vary node called $VN$.
>> For $k$ := $1, 2, \ldots n_i$
>>> Let $Childnums_k$ := $\{\}$.
>> For each $j \in RecordNums$
>>> Let $v_{ij}$ = Value of attribute $a_i$ in record $j$
>>> Add $j$ to the set $Childnums_{v_{ij}}$
>> Let $VN.MCV$ := $\text{argmax}_k | Childnums_k |$.
>> For $k$ := $1, 2, \ldots n_i$
>>> If $| Childnums_k | = 0$ or if $k = MCV$
>>>> Set the $a_i = k$ subtree of $VN$ to NULL.
>>> Else
>>>> Set the $a_i = k$ subtree of $VN$ to **MakeADTree**($a_{i+1}, Childnums_k$)

To build the entire tree, we must call **MakeADTree**($a_1, \{1 \ldots R\}$). Assuming binary attributes, the cost of building a tree from $R$ records and $M$ attributes is bounded above by

$$\sum_{k=0}^{\lfloor \log_2 R \rfloor} \frac{R}{2^k} \begin{pmatrix} M \\ k \end{pmatrix} \tag{18}$$